\documentclass[11pt,letterpaper]{article}
\usepackage{emnlp2016}
\usepackage{times}
\usepackage{latexsym}
\usepackage[T1]{fontenc}

\usepackage{times}
\usepackage{latexsym}
\usepackage{url}
\usepackage{floatrow}
\usepackage{xcolor}
\usepackage{amsmath, graphicx}
\usepackage{amsfonts, amssymb}
\usepackage{algorithmicx}
\usepackage{algorithm}

\usepackage[noend]{algpseudocode}
\usepackage[font={small},labelfont=bf]{caption}
\usepackage{breqn}
\usepackage{multirow, array} 
\usepackage{tikz-dependency}

\usepackage{tabularx}
\newcolumntype{Y}{>{\centering\arraybackslash}X}

\usepackage{array}
\newcolumntype{L}{>{\centering\arraybackslash}m{3cm}}

 \usepackage[linewidth=1pt]{mdframed}
 
\usepackage{floatrow}%
\usepackage{arydshln}

\usepackage{enumitem}
\setlist{nolistsep,leftmargin=*} 

\usepackage[draft, backgroundcolor=orange!30]{todonotes}   

 \newcommand{\comment}[1]{}  

\emnlpfinalcopy 
\def\emnlppaperid{559} 

\NewEnviron{elaboration}{
\par
\begin{tikzpicture}
\node[rectangle,minimum width=\textwidth] (m) {\begin{minipage}{0.98\textwidth}\BODY\end{minipage}};
\draw[dashed] (m.south west) rectangle (m.north east);
\end{tikzpicture}
}

\title{Improving Information Extraction by Acquiring External Evidence with Reinforcement Learning} 

\author{Karthik Narasimhan\\
	    CSAIL, MIT\\
	    {\tt karthikn@mit.edu}
	  \And
	  Adam Yala\\
	    CSAIL, MIT\\
	    {\tt adamyala@mit.edu}
	   \And
	   Regina Barzilay\\
	    CSAIL, MIT\\
	    {\tt regina@csail.mit.edu}
	}
\date{}

\begin{document}
\maketitle
\begin{abstract}

Most successful information extraction systems operate with access to a large collection of documents. In this work, we explore the task of acquiring and incorporating external evidence to improve extraction accuracy in domains where the amount of training data is scarce. This process entails issuing search queries, extraction from new sources and reconciliation of extracted values, which are repeated until sufficient evidence is collected. We approach the problem using a reinforcement learning framework where our model learns to select optimal actions based on contextual information. We employ a deep Q-network, trained to optimize a reward function that reflects extraction accuracy while penalizing extra effort. Our experiments on two databases -- of shooting incidents, and food adulteration cases -- demonstrate that our system significantly outperforms traditional extractors and a competitive meta-classifier baseline.\footnote{Code is available at \url{http://people.csail.mit.edu/karthikn/rl-ie/}}

\end{abstract}
\section{Introduction}
\label{sec:introduction}

\begin{figure}[t]
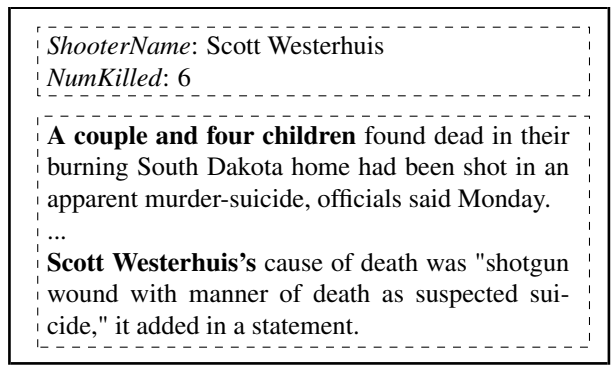

\small
\begin{mdframed}
\begin{elaboration}
\emph{ShooterName}: Scott Westerhuis \\
\emph{NumKilled}: 6 
\end{elaboration}
\vspace{-0.2cm}
\begin{elaboration}
  \parbox{0.98\textwidth}{
\textbf{A couple and four children} found dead in their burning South Dakota home had been shot in an apparent murder-suicide, officials said Monday.

 ... 

 \textbf{Scott Westerhuis's} cause of death was "shotgun wound with manner of death as suspected suicide," it added in a statement.}
\end{elaboration}

\end{mdframed}

\caption{Sample news article on a shooting case. Note how the article contains both the name of the shooter and the number of people killed but both pieces of information require complex extraction schemes.}
\label{example1}
\end{figure}

In many realistic domains, information extraction (IE) systems require exceedingly large amounts of annotated data to deliver high performance. Increases in training data size enable models to handle robustly the multitude of linguistic expressions that convey the same semantic relation.  Consider, for instance, an IE system that aims to identify entities such as the perpetrator and the number of victims in a shooting incident (Figure~\ref{example1}).
The document does not explicitly mention the shooter (\emph{Scott Westerhuis}), but instead refers to him as a suicide victim. Extraction of the number of fatally shot victims is similarly difficult, as the system needs to infer that \emph{"A couple and four children"} means \emph{six people}.  Even a large annotated training set may not provide sufficient coverage to capture such challenging cases.


In this paper, we explore an alternative approach for boosting extraction accuracy, when a large training corpus is not available.  Instead, the proposed method utilizes external information sources to resolve ambiguities inherent in text interpretation. Specifically, our strategy is to find other documents that contain the information sought, expressed in a form that a basic extractor can "understand". For instance, Figure~\ref{example2} shows two other articles describing the same event, wherein the entities of interest -- the number of people killed and the name of the shooter -- are expressed explicitly. Processing such stereotypical phrasing is easier for most extraction systems, compared to analyzing the original source document. This approach is particularly suitable for extracting information from news where a typical event is covered by multiple news outlets. 

The challenges, however, lie in (1) performing \emph{event coreference} (i.e. retrieving suitable articles describing the same incident) and (2) reconciling the entities extracted from these different documents. Querying the web (using the source article's title for instance) often retrieves documents about other incidents with a tangential relation to the original story. For example, the query ``\emph{4 adults, 1 teenager shot in west Baltimore 3 april 2015}'' yields only two relevant articles among the top twenty results on Bing search, while returning other shooting events at the same location. Moreover, the values extracted from these different sources require resolution since some of them might be inaccurate.


\begin{figure}[t]
\small
\begin{mdframed}
\begin{elaboration}
  \parbox{0.98\textwidth}{
The \textbf{six} members of a South Dakota family found dead in the ruins of their burned home were fatally shot, with one death believed to be a suicide, authorities said Monday.
}
\end{elaboration}
\vspace{-0.2cm}
\begin{elaboration}
  \parbox{0.98\textwidth}{
 AG Jackley says all evidence supports the story he told based on preliminary findings back in September: \textbf{Scott Westerhuis} shot his wife and children with a shotgun, lit his house on fire with an accelerant, then shot himself with his shotgun.
}
\end{elaboration}
\end{mdframed}

\caption{Two other articles on the same shooting case. The first article clearly mentions that six people were killed. The second one portrays the shooter in an easily extractable form.}
\label{example2}
\end{figure}

One solution to this problem would be to perform a single search to retrieve articles on the same event and then reconcile values extracted from them (say,  using a \emph{meta-classifier}).
  However, if the confidence of the new set of values is still low, we might wish to perform further queries.
Thus, the problem is inherently sequential, requiring alternating phases of querying to retrieve articles and integrating the extracted values.

We address these challenges using a Reinforcement Learning (RL) approach that combines query formulation, extraction from new sources, and value reconciliation. To effectively select among possible actions, our state representation encodes information about the current and new entity values along with the similarity between the source article and the newly retrieved document. The model learns to select good actions for both article retrieval and value reconciliation in order to optimize the reward function, which reflects extraction accuracy and includes penalties for extra moves. 
 We train the RL agent using a Deep Q-Network (DQN)~\cite{mnih2015dqn} that is used to predict both querying and reconciliation choices simultaneously. While we use a maximum entropy model as the base extractor, this framework can be inherently applied to other extraction algorithms.

We evaluate our system on two datasets where available training data is inherently limited.  The first dataset is constructed from a publicly available database of mass shootings in the United States. The database is populated by volunteers and includes the source articles.  The second dataset is derived from a FoodShield database of illegal food adulterations.  Our experiments demonstrate that the final RL model outperforms basic extractors as well as a meta-classifier baseline in both domains. For instance, in the \emph{Shootings} domain, the average accuracy improvement over the meta-classifier is 7\%.

\section{Related Work}
\label{sec:relatedwork}

\paragraph{Open Information Extraction} Existing work in open IE has used external sources from the web to improve extraction accuracy and coverage~\cite{agichtein2000snowball,etzioni2011open,fader2011identifying,wu2010open}. Such research has focused on identifying multiple instances of the same relation, independent of the context in which this information appears. In contrast, our goal is to extract information from additional sources about a specific event described in a source article. Therefore, the novel challenge of our task resides in performing event coreference~\cite{lee2012eventcoref,bejan2014unsupervised} (i.e identifying other sources describing the same event) while simultaneously reconciling extracted information. Moreover, relations typically considered by open IE systems have significantly higher coverage in online documents than a specific incident described in a few news sources. Hence, we require a different mechanism for finding and reconciling online information.

\paragraph{Entity linking, multi-document extraction and event coreference} Our work also relates to the task of multi-document information extraction, where the goal is to connect different mentions of the same entity across input documents~\cite{mann2005multi,han2011collective,durrett2014joint}. Since this setup already includes multiple input documents, the model is not required to look for additional sources or decide on their relevance. Also, while the set of input documents overlap in terms of entities mentioned, they do not necessarily describe the same event.  Given these differences in setup, the challenges and opportunities of the two tasks are distinct.


\paragraph{Knowledge Base Completion and Online Search}
Recent work has explored several techniques to perform Knowledge Base Completion (KBC) such as vector space models and graph traversal~\cite{socher2013reasoning,yang2014embedding,gardner2014kbc,neelakantan2015compositional,guu2015traversing}. 
Though our work also aims at increasing extraction recall for a database, traditional KBC approaches do not require searching for additional sources of information. \newcite{west2014knowledge} explore query reformulation in the context of KBC. Using existing search logs, they learn how to formulate effective queries for different types of database entries. Once query learning is completed, the model employs several selected queries, and then aggregates the results based on retrieval ranking. This approach is complementary to the proposed method, and can be combined with our approach if search logs are available.

\newcite{kanani2012selecting} also combine search and information extraction. In their task of faculty directory completion, the system has to find documents from which to extract desired information. They employ reinforcement learning to address computational bottlenecks, by minimizing the number of queries, document downloads and extraction action. The extraction accuracy is not part of this optimization, since the baseline IE system achieves high performance on the relations of interest. Hence, given different design goals, the two RL formulations are very different.
Our approach is also close in spirit to the AskMSR system~\cite{banko2002askmsr} which aims at using information redundancy on the web to better answer questions. Though our goal is similar, we learn to query and consolidate the different sources of information instead of using pre-defined rules.
Several slot-filling methods have experimented with query formulation over web-based corpora to populate knowledge bases~\cite{surdeanu2010simple,ji2011kbp}.

\section{Framework}
\label{sec:framework}

\begin{figure*}

\begin{minipage}{.64\linewidth}
  \includegraphics[width=\linewidth]{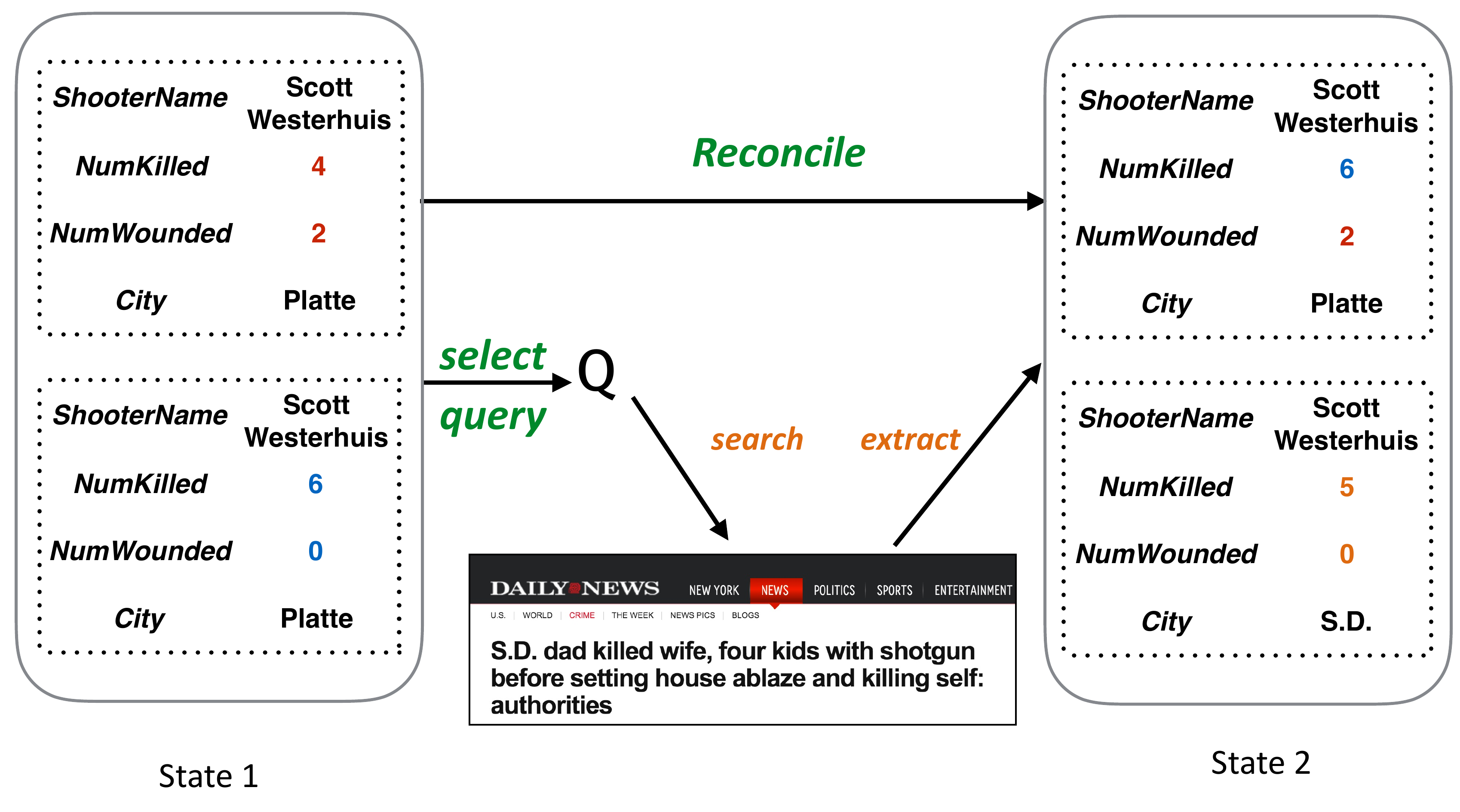}
\end{minipage}
\resizebox{.35\linewidth}{!}{
\begin{minipage}{0.5\linewidth}

	\begin{mdframed}
	\textbf{Current Values:} \\
	\emph{ShooterName} $\rightarrow$ Scott Westerhuis \\
	\emph{NumKilled} $\rightarrow$ 4 \\
	\emph{NumWounded} $\rightarrow$ 2 \\
	\emph{City} $\rightarrow$ Platte  \vspace{0.2cm} \\
	\textbf{New Values:} \\
	\emph{ShooterName} $\rightarrow$ Scott Westerhuis \\
	\emph{NumKilled} $\rightarrow$ 6 \\
	\emph{NumWounded} $\rightarrow$ 0 \\
	\emph{City} $\rightarrow$ Platte
	\end{mdframed}
	\begin{mdframed}
	\textbf{State:} \\
	 $\langle 0.3, 0.2, 0.5, 0.1, \hfill \leftarrow \text{currentConf} \\ 
	 0.4, 0.6, 0.2, 0.4, \hfill \leftarrow \text{newConf} \\ 
	 1, 0, 0, 1, 0, 1, 1, 0, \hfill \leftarrow \text{matches}  \\ 
	 0.2, 0.3, ... , 0.1, 0.5,  \hfill \leftarrow \text{contextWords} \\
	 0.65\rangle \hfill \leftarrow \text{document tf-idf similarity}  $
	\end{mdframed}
\end{minipage}%
}
\caption{ \textbf{Left: } Illustration of a transition in the MDP -- the top box in each state shows the current entities and the bottom one consists of the new entities extracted from a downloaded article on the same event. \textbf{Right:} Sample state representation (bottom) in the MDP based on current and new values of entities (top). \emph{currentConf}: confidence scores of current entities, \emph{newConf}: confidence scores of new entities, \emph{contextWords}: tf-idf counts of context words.}
\label{fig:mdp}
\end{figure*}

We model the information extraction task as a markov decision process (MDP), where the model learns to utilize external sources to improve upon extractions from a source article (see Figure~\ref{fig:mdp}). The MDP framework allows us to dynamically incorporate entity predictions while also providing flexibility to choose the type of articles to extract from. At each step, the system has to reconcile extracted values  from a related article ($e_{new}$) with the current set of values ($e_{cur}$), and decide on the next query for retrieving more articles.

We represent the MDP as a tuple $\langle S, A, T, R \rangle$, where $S=\{s\}$ is the space of all possible states, $A = \{a = (d,q)\}$  is the set 
of all actions, $R(s, a)$ is the reward function, and  $T(s' | s, a)$ is the transition function. We describe these in detail below.


\paragraph{States}  
The state $s$ in our MDP consists of the extractor's confidence in predicted entity values, the context from which the values are extracted and the similarity between the new document and the original one. We represent the state as a continuous real-valued vector (Figure~\ref{fig:mdp}) incorporating these pieces of information:
\begin{enumerate}
\item Confidence scores of current and newly extracted entity values.
\item One-hot encoding of matches between current and new values. 
\item Unigram/tf-idf counts\footnote{Counts are computed on the documents used to train the basic extraction system.} of context words. These are words that occur in the neighborhood of the entity values in a document (e.g. the words \emph{which}, \emph{left}, \emph{people} and \emph{wounded} in the phrase ``\emph{which left 5 people wounded}'').
\item \emph{tf-idf} similarity between the original article and the new article.
\end{enumerate}

 \paragraph{Actions}
 At each step, the agent is required to take two actions - a reconciliation decision $d$ and a query choice $q$. The decision $d$ on the newly extracted values can be one of the following types: (1) accept a specific entity's value (one action per entity)\footnote{No entity-specific features are used for action selection.}, (2) accept all entity values, (3) reject all values or (4) stop. In cases 1-3, the agent continues to inspect more articles, while the episode ends if a stop action (4) is chosen. The current values and confidence scores are simply updated with the accepted values and the corresponding confidences.\footnote{We also experiment with other forms of value reconciliation. See Section~\ref{sec:experiments} for details.} The choice $q$ is used to choose the next query from a set of automatically generated alternatives (details below) in order to retrieve the next article. 
 

 \paragraph{Rewards}
 The reward function is chosen to maximize the final \emph{extraction accuracy} while minimizing the number of queries. The accuracy component is calculated using the difference between the accuracy of the current and the previous set of entity values:
 \begin{dmath*}
  R(s, a) 	= \sum_{\text{entity}~j} \text{Acc}(e_{cur}^j) - \text{Acc}(e_{prev}^j)
 \end{dmath*}
There is a negative reward per step to penalize the agent for longer episodes. 


\paragraph{Queries}
The queries are based on automatically generated templates, created using the title of an article along with words\footnote{Stop words, numeric terms and proper nouns are filtered.} most likely to co-occur with each entity type in the training data. Table~\ref{table:query-types} provides some examples -- for instance, the second template contains words such as \emph{arrested} and \emph{identified} which often appear around the name of the shooter.

 \begin{table}[!htbp]
\centering
\resizebox{\textwidth}{!}{%
\begin{tabular}{|| c ||} \hline
\emph{$\langle$title$\rangle$} \\
\emph{$\langle$title$\rangle$} + (police | identified | arrested | charged) \\
\emph{$\langle$title$\rangle$} + (killed | shooting | injured | dead | people) \\
\emph{$\langle$title$\rangle$} + (injured | wounded | victim)\\
\emph{$\langle$title$\rangle$} + (city | county | area) \\ \hline
\end{tabular}
}
\caption{Examples of different query templates for web search for articles on mass shootings. The $|$ symbol represents logical OR. The last 4 queries contain context words around values for entity types ShooterName, NumKilled, NumWounded and City, respectively. At query time, \emph{$\langle$title$\rangle$} is replaced by the source article's title. }
\label{table:query-types}
\end{table}

 We use a search engine to query the web for articles on the same event as the source article and retrieve the top $k$ links per query.\footnote{We use $k$=20 in our experiments.} Documents that are more than a month older than the original article are filtered out of the search results.

\paragraph{Transitions}
Each episode starts off with a single source article $x_i$ from which an initial set of entity values are extracted. The subsequent steps in the episode involve the extra articles, downloaded using different types of query formulations based on the source article. A single transition in the episode consists of the agent being given the state $s$ containing information about the current and new set of values (extracted from a single article) using which the next action $a=(d,q)$ is chosen.   The transition function $T(s' | s, a)$ incorporates the reconciliation decision $d$ from the agent in state $s$ along with the values from the next article retrieved using query $q$ and produces the next state $s'$. The episode stops whenever $d$ is a stop decision.

Algorithm~\ref{alg:server} details the entire MDP framework for the training phase. During the test phase, each source article is handled only once in a single episode (lines 8-23).

\begin{algorithm}
\small
\caption{\small MDP framework for Information Extraction (Training Phase)}
\label{alg:server}

\begin{algorithmic}[1]
\State Initialize set of original articles $X$
\For {$ x_i \in X $}
	\For { each query template $T^q $}
		\State Download articles with query $T^q(x_i)$
		\State Queue retrieved articles in $Y_i^q$
	\EndFor
\EndFor
\For {$ epoch = 1,M $}
	\For {$ i = 1, |X| $}  \hspace{1cm}   //episode
		\State Extract entities $e_0$ from $x_i$
		\State $e_{cur}$ $\leftarrow$ $e_0$
		\State q  $\leftarrow$ 0, r $\leftarrow$ 0  \hspace{0.5cm} //query type, reward
		\While { $Y_i^q$ not empty }
			\State Pop next article $y$ from $Y_i^q$
			\State Extract entities $e_{new}$ from $y$
			\State Compute tf-idf similarity $\mathcal{Z}(x_i, y)$
			\State Compute context vector $\mathcal{C}(y)$
			\State  Form state $s$ using $e_{cur}$, $e_{new}$, $\mathcal{Z}(x_i, y)$ and  $\mathcal{C}(y)$
			\State Send ($s$, $r$) to agent
			\State Get decision $d$, query $q$ from agent
			\If { $q$ == ``end_episode''}
				\textbf{break}
			\EndIf
			\State $e_{prev}$ $\leftarrow$ $e_{cur}$
			\State $e_{cur}$ $\leftarrow$ \emph{Reconcile}($e_{cur}$, $e_{new}$, $d$)
%
		 \State $r \leftarrow \sum_{\text{entity}~j} \text{Acc}(e_{cur}^j) - \text{Acc}(e_{prev}^j)$ 
		 \EndWhile
		\State Send ($s_{end}$, $r$) to agent

	\EndFor
\EndFor
\end{algorithmic}
\normalsize
\end{algorithm}

\section{Reinforcement Learning for Information Extraction}
\label{sec:model}

In order to learn a good policy for an agent, we utilize the paradigm of reinforcement learning (RL). The MDP described in the previous section
can be viewed in terms of a sequence of transitions ($s, a, r, s'$). The agent typically utilizes a state-action value function $Q(s,a)$ to determine which action $a$ to perform in state $s$. A commonly used technique for learning an optimal value function is Q-learning~\cite{watkins1992q}, in which the agent iteratively updates $Q(s,a)$ using the rewards obtained from episodes. The updates are derived from the recursive Bellman equation~\cite{sutton1998introduction} for the optimal Q:
\begin{dmath*}
	{Q_{i+1}(s,a) = \mathrm{E}[r + \gamma \max_{a'} Q_i(s',a') \mid s, a]}
\label{eq:bellman-update}
\end{dmath*}
Here,  $r = R(s,a)$ is the reward and $\gamma$ is a factor discounting the value of future rewards and the expectation is taken over all transitions involving state $s$ and action $a$. 

Since our problem involves a continuous state space $S$, we use a deep Q-network (DQN)~\cite{mnih2015dqn} as a function approximator ${Q(s,a) \approx Q(s, a; \theta)}$.
The DQN, in which the Q-function is approximated using a deep neural network, has been shown to learn better value functions than linear approximators~\cite{narasimhan2015language,he2015deep} and can capture non-linear interactions between the different pieces of information in our state.


We use a DQN consisting of two linear layers (20 hidden units each)  followed by rectified linear units (ReLU), along with two separate output layers.\footnote{We did not observe significant differences with additional linear layers or the choice of non-linearity (Sigmoid/ReLU).}  The network takes the continuous state vector $s$ as input and predicts $Q(s, d)$ and $Q(s, q)$ for reconciliation decisions $d$ and query choices $q$ simultaneously using the different output layers (see Supplementary material for the model architecture).


 \begin{algorithm}
\small
\caption{\small Training Procedure for DQN agent with $\epsilon$-greedy exploration}
\label{alg:training}

\begin{algorithmic}[1]
\State Initialize experience memory $\mathcal{D}$
\State Initialize parameters $\theta$ randomly
\For {$ episode = 1,M $}
	\State Initialize environment and get start state $s_1$	
	\For {$ t = 1, N $}
		\If {$ random() < \epsilon $}			
			\State Select a random action $a_t$
		\Else		
			\State Compute $Q(s_t, a)$ for all actions $a$
			\State Select  $a_t = \text{argmax}~Q(s_t, a)$
		\EndIf
		\State Execute action $a_t$ and observe reward $r_t$ and new state $s_{t+1}$
		\State Store transition $(s_t, a_t, r_t, s_{t+1})$ in $\mathcal{D}$
		\State Sample random mini batch of transitions $(s_j, a_j, r_j, s_{j+1})$ from $\mathcal{D}$
		\State {$y_j = \left\{
		\begin{array}{lcl}
 				       r_j ,~~~~~~~~~~~~~~~~~~~~~~~~\text{if } s_{j+1} \text{ is terminal}\\
				       r_j + \gamma~\text{max}_{a'}~Q(s_{j+1}, a'; \theta_{t}),~~~\text{else} \\ 
	      \end{array}
		    \right. $}
		\State Perform gradient descent step on the loss $\mathcal{L}(\theta) = (y_j - Q(s_j, a_j; \theta))^2$ 
		\If { $s_{t+1}$ == $s_{end}$}
			\textbf{break}
		\EndIf
	\EndFor
\EndFor

\end{algorithmic}
\normalsize
\end{algorithm}

\paragraph{Parameter Learning}
The parameters $\theta$ of the DQN are learnt using stochastic gradient descent with RMSprop~\cite{tieleman2012lecture}. Each parameter update aims to close the gap between the $Q(s_t, a_t; \theta)$ predicted by the DQN and the expected Q-value from the Bellman equation, ${r_t + \gamma \max_a Q(s_{t+1}, a; \theta)}$. Following \newcite{mnih2015dqn}, we make use of a (separate) target Q-network to calculate the expected Q-value, in order to have `stable updates'. 
The target Q-network is periodically updated with the current parameters $\theta$.
We also make use of an experience replay memory $\mathcal{D}$ to store transitions. To perform updates, we sample a batch of transitions ($\hat{s}, \hat{a}, \hat{s}', r$) at random from $\mathcal{D}$ and minimize the loss function\footnote{The expectation is over the transitions sampled uniformly at random from $\mathcal{D}$.}:
\begin{dmath*}
	{\mathcal{L}(\theta) = \mathrm{E}_{\hat{s},\hat{a}}  [ (y - Q(\hat{s}, \hat{a} ; \theta))^2 ]}
\label{eq:loss}
\end{dmath*}
where $ {y = r + \gamma \max_{a'} Q (\hat{s}',a'; \theta_{t}) }$ is the target Q-value. The learning updates are made every training step using the following gradients:
\begin{dmath*}
	{\nabla_{\theta} \mathcal{L}(\theta) = \mathrm{E}_{\hat{s},\hat{a}}  [ 2(y - Q(\hat{s}, \hat{a} ; \theta)) \nabla_{\theta} Q(\hat{s}, \hat{a} ; \theta) ]} 
\label{eq:loss-gradient}
\end{dmath*}
 Algorithm~\ref{alg:training} details the DQN training procedure.


\section{Experimental Setup}
\label{sec:experiments} 

\paragraph{Data}
We perform experiments on two different datasets. For the first set, we collected data from the Gun Violence archive,\footnote{\url{www.shootingtracker.com/Main_Page}} a website tracking shootings in the United States. The data contains a news article on each shooting and annotations for (1) the name of the shooter, (2) the number of people killed, (3) the number of people wounded, and (4) the city where the incident took place. We consider these as the entities of interest, to be extracted from the articles.  The second dataset we use is the Foodshield EMA database\footnote{\url{www.foodshield.org/member/login/}} documenting  adulteration incidents since 1980. This data contains annotations for (1) the affected food product, (2) the adulterant and (3) the location of the incident. Both datasets are classic examples where the number of recorded incidents is insufficient for large-scale IE systems to leverage.

For each source article in the above databases, we download extra articles (top 20 links) using the Bing Search API\footnote{\url{www.bing.com/toolbox/bingsearchapi}}  with  different automatically generated queries.
We use only the source articles from the \emph{train} portion to learn the parameters of the base extractor. The entire \emph{train} set with downloaded articles is used to train the DQN agent and the meta-classifier baseline (described below). All parameters are tuned on the \emph{dev} set. For the final results, we train the models on the combined train and dev sets and use the entire \emph{test} set (source + downloaded articles) to evaluate. Table~\ref{table:data} provides data statistics.


\begin{table}[t]
\small
\centering
\resizebox{\textwidth}{!}{%
\begin{tabular}{ c | c | c | c | c | c | c } 
\multirow{2}{*}{Number} &  \multicolumn{3}{ c |}{Shootings} & \multicolumn{3}{ c }{Adulteration}\\ 
 & \textbf{Train} & \textbf{Test} & \textbf{Dev} & \textbf{Train} & \textbf{Test} & \textbf{Dev} \\ \hline 
Source articles & 306 & 292 & 66 & 292 & 148 & 42 \\
Downloaded articles & 8201 & 7904 & 1628 & 7686 & 5333 & 1537 \\
\end{tabular}
}
\caption{Stats for \emph{Shootings} and \emph{Adulteration} datasets}
\label{table:data}
\end{table}



\paragraph{Extraction model}
We use a maximum entropy classifier as the base extraction system, since it provides flexibility to capture various local context features and has been shown to perform well for information extraction~\cite{chieu2002maximum}. The classifier is used to tag each word in a document as one of the entity types or not (e.g. \{\emph{ShooterName, NumKilled, NumWounded, City, Other}\} in the \emph{Shootings} domain). Then, for each tag except \emph{Other}, we choose the mode of the values to obtain the set of entity extractions from the article.\footnote{We normalize numerical words (e.g. "one" to "1") before taking the mode.} Features used in the classifier are provided in the Supplementary material.

The features and context window $c=4$ of neighboring words are tuned to maximize performance on a dev set. We also experimented with a conditional random field (CRF) (with the same features) for the sequence tagging~\cite{culotta2004confidence} but obtained worse empirical performance (see Section~\ref{sec:results}). The parameters of the base extraction model are not changed during training of the RL model.

\paragraph{Evaluation}
We evaluate the extracted entity values against the gold annotations and report the corpus-level average accuracy on each entity type. For entities like ShooterName, the annotations (and the news articles) often contain multiple names (first and last) in various combinations, so we consider retrieving either name as a successful extraction. For all other entities, we look for exact matches.

\paragraph{Baselines}
We explore 4 types of baselines: 

\emph{Basic extractors:} We use the CRF and the Maxent classifier mentioned previously.

\emph{Aggregation systems:} We examine two systems that perform different types of value reconciliation. The first model (\emph{Confidence}) chooses entity values with the highest confidence score assigned by the base extractor. The second system (\emph{Majority}) takes a majority vote over all values extracted from these articles. Both methods filter new entity values using a threshold $\tau$ on the cosine similarity over the tf-idf representations of the source and new articles. 

\emph{Meta-classifer:} To demonstrate the importance of modeling the problem in the RL framework, we consider a meta-classifier baseline. The classifier operates over the same input state space and produces the same set of reconciliation decisions $\{d\}$ as the DQN. For training, we use the original source article for each event along with a related downloaded article to compute the state. If the downloaded article has the correct value and the original one does not, we label it as a positive example for that entity class. If multiple such entity classes exist, we create several training instances with appropriate labels, and if none exist, we use the label corresponding to the \emph{reject all} action. For each \emph{test} event, the classifier is used to provide decisions for all the downloaded articles and the final extraction is performed by aggregating the value predictions using the \emph{Confidence}-based scheme described above. 

\emph{Oracle:} Finally, we also have an \textsc{Oracle} score which is computed assuming perfect reconciliation and querying decisions on top of the Maxent base extractor. This helps us analyze the contribution of the RL system in isolation of the inherent limitations of the base extractor.

\paragraph{RL models}
We perform experiments using three variants of RL agents -- (1) \emph{RL-Basic}, which performs only reconciliation decisions\footnote{Articles are presented to the agent in a round-robin fashion from the different query lists.}, (2) \emph{RL-Query}, which takes only query decisions with the reconciliation strategy fixed (similar to \newcite{kanani2012selecting}), and  (3) \emph{RL-Extract}, our full system incorporating both reconciliation and query decisions.

We train the models for 10000 steps every epoch using the Maxent classifier as the base extractor, and evaluate on the entire \emph{test} set every epoch. 
The final accuracies reported are averaged over 3 independent runs; each run's score is averaged over 20 epochs after 100 epochs of training. The penalty per step is set to -0.001. 
For the DQN, we use the dev set to tune all parameters. We used a replay memory $\mathcal{D}$ of size 500k, and a discount ($\gamma$) of 0.8. We set the learning rate to $2.5\text{E}^{-5}$. The $\epsilon$ in $\epsilon$-greedy exploration is annealed from 1 to 0.1 over 500k transitions. The target-Q network is updated every 5k steps.



%
%
%
 

\section{Results}
\label{sec:results}

\paragraph{Performance}
Table~\ref{table:results} demonstrates that our system (RL-Extract) obtains a substantial gain in accuracy over the basic extractors on all entity types over both domains. For instance, RL-Extract is 11.4\% more accurate than the basic Maxent extractor on City and 7.1\% better on NumKilled, while also achieving gains of more than 5\% on the other entities on the \emph{Shootings} domain. The gains on the \emph{Adulteration} dataset are also significant, up to a 11.5\% increase on the Location entity.

\begin{table*}[!htbp]
\centering
\resizebox{\textwidth}{!}{%
\begin{tabular}{ c  || c | c | c | c || c | c | c  } 
 \multirow{ 2}{*}{\textbf{System}} &  \multicolumn{4} { c || }{\textbf{Shootings}} &  \multicolumn{3} { c }{\textbf{Adulteration}} \\ 
&  \textbf{ShooterName} & \textbf{NumKilled} & \textbf{NumWounded} & \textbf{City}  &  \textbf{Food} & \textbf{Adulterant} & \textbf{Location}  \\ \hline 
 \emph{CRF extractor} & 9.5 & 65.4 & 64.5 & 47.9 & 41.2 & 28.3 & 51.7 \\
 \emph{Maxent extractor} & 45.2 & 69.7 & 68.6 & 53.7 & 56.0 & 52.7 & 67.8\\ \hdashline[0.5pt/5pt]
 \emph{Confidence Agg. ($\tau$)}  & 45.2 (0.6) & 70.3 (0.6) & 72.3 (0.6) & 55.8 (0.6) & 56.0 (0.8) & 54.0 (0.8) & 69.2 (0.6) \\
 \emph{Majority Agg. ($\tau$)}  & 47.6 (0.6) & 69.1 (0.9) & 68.6 (0.9) & 54.7 (0.7) & 56.7 (0.5) & 50.6 (0.95) & 72.0 (0.4) \\  \hdashline[0.5pt/5pt]
  \emph{Meta-classifier} & 45.2 & 70.7 & 68.4 & 55.3 & 55.4 & 52.7 & 72.0 \\ \hline
  RL-Basic & 45.2 & 71.2 & 70.1 & 54.0 & 57.0 & 55.1 & 76.1\\
  RL-Query (conf) &  39.6 & 66.6 & 69.4 & 44.4 & 39.4 & 35.9 & 66.4\\
  RL-Extract & \textbf{50.0} & \textbf{77.6}$^*$ & \textbf{74.6}$^*$ & \textbf{65.6}$^*$ & \textbf{59.6}$^*$ & \textbf{58.9}$^*$ & \textbf{79.3}$^*$\\  \hline
  \textsc{Oracle} & 57.1 & 86.4 & 83.3 & 71.8 & 64.8 & 60.8 & 83.9 \\ 
\end{tabular}

}
\caption{Accuracy of various baselines (italics), our system (DQN) and the Oracle on \emph{Shootings} and \emph{Adulteration} datasets.
\textbf{Agg.}~refers to aggregation baselines. Bold indicates best system scores. $^{*}$statistical significance  of $p < 0.0005$   vs basic Maxent extractor using the Student-t test. Numbers in parentheses indicate the optimal threshold ($\tau$) for the aggregation baselines. Confidence-based reconciliation was used for RL-Query.}
\label{table:results}
\end{table*}

\begin{figure}[!hbp]
  \includegraphics[width=0.95\linewidth]{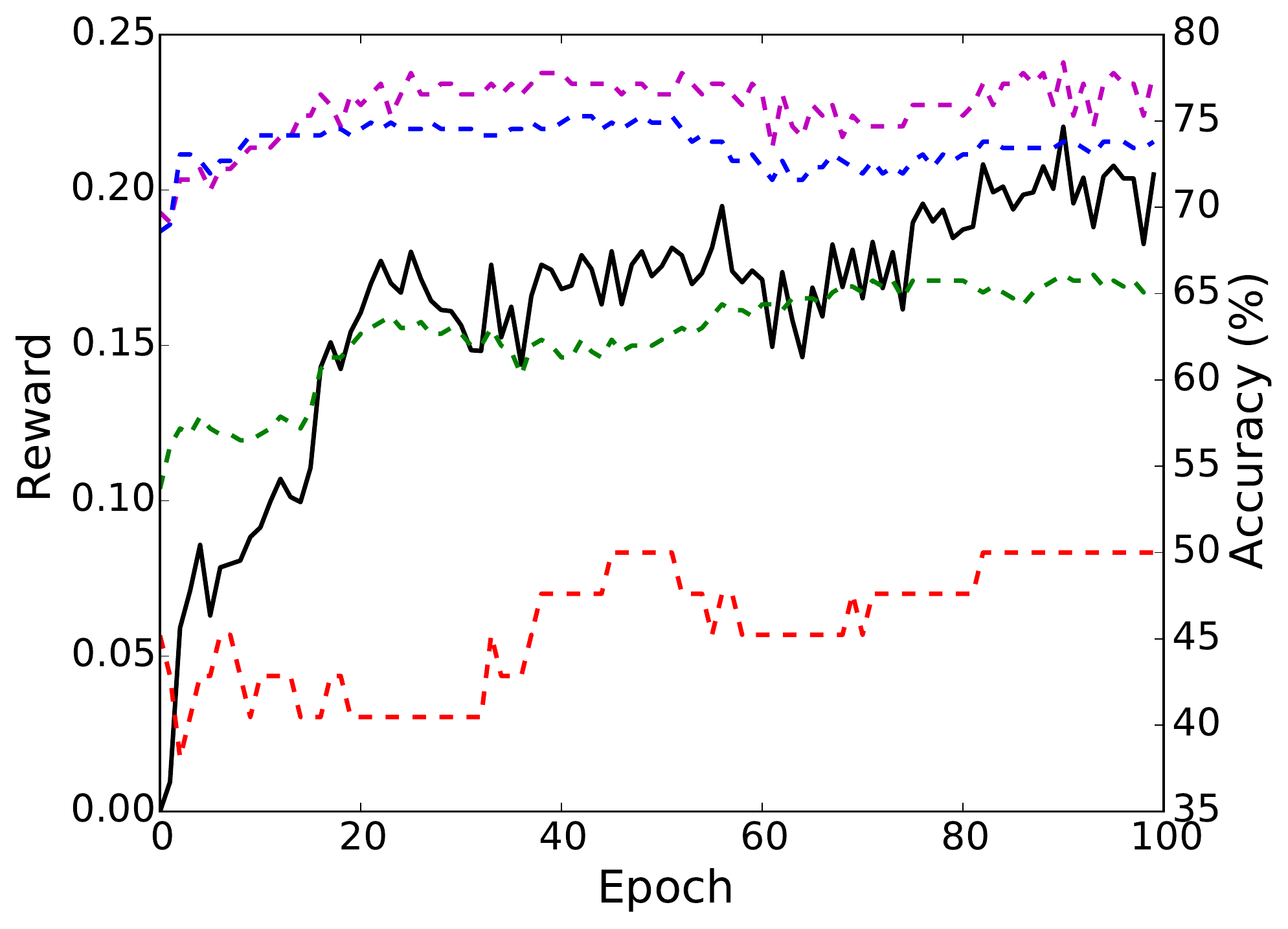}
\caption{Evolution of average reward (solid black) and accuracy on various entities (dashed lines; red=\emph{ShooterName}, magenta=\emph{NumKilled}, blue=\emph{NumWounded}, green=\emph{City}) on the \emph{test} set of the \emph{Shootings} domain.}
	\label{fig:reward}
\end{figure}

 \begin{table*}[t]
 \footnotesize
\centering
\begin{tabular}{ | c | c | L | } \hline 
\textbf{Entity} & \textbf{System: Value} & \textbf{Example} \\ \hline
\multirow{2}{*}{ShooterName} & \textbf{Basic}: Stewart & \multicolumn{1}{m{10cm}|}{A source tells Channel 2 Action News that Thomas Lee has been arrested in Mississippi ...  Sgt . Stewart Smith, with the Troup County Sheriff's office, said. } \\ \cdashline{2-3}[0.5pt/5pt]
& \textbf{RL-Extract}: Lee &  \multicolumn{1}{m{10cm}|}{Lee is accused
       of killing his wife, Christie; ...} \\ \hline 
\multirow{2}{*}{NumKilled} & \textbf{Basic}: 0 & \multicolumn{1}{m{10cm}|}{Shooting leaves 25 year old Pittsfield man dead , 4 injured} \\ \cdashline{2-3}[0.5pt/5pt]
& \textbf{RL-Extract}: 1 &  \multicolumn{1}{m{10cm}|}{One man is dead after a shooting Saturday night at the intersection of Dewey Avenue and Linden Street.} \\ \hline 
\multirow{2}{*}{NumWounded} & \textbf{Basic}: 0 & \multicolumn{1}{m{10cm}|}{Three people are dead and a fourth is in the hospital after a murder suicide} \\ \cdashline{2-3}[0.5pt/5pt]
& \textbf{RL-Extract}: 1 &  \multicolumn{1}{m{10cm}|}{3 dead, 1 injured in possible Fla. murder-suicide} \\ \hline 
\multirow{2}{*}{City} & \textbf{Basic}: Englewood & \multicolumn{1}{m{10cm}|}{A 2 year old girl and four other people were wounded in a shooting in West Englewood Thursday night, police said} \\  \cdashline{2-3}[0.5pt/5pt]
& \textbf{RL-Extract}: Chicago &  \multicolumn{1}{m{10cm}|}{At least 14 people were shot across Chicago between noon and 10:30 p.m. Thursday. The last shooting left five people wounded.} \\ \hline 
\end{tabular}
\caption{Sample outputs (along with corresponding article snippets) on the \emph{Shootings} domain showing correct predictions from RL-Extract where the basic extractor (Maxent) fails.}
\label{table:outputs}
\end{table*}

We can also observe that simple aggregation schemes like the \emph{Confidence} and \emph{Majority} baselines don't handle the complexity of the task well. RL-Extract outperforms these by 7.2\% on \emph{Shootings} and 5\% on \emph{Adulteration} averaged over all entities. Further, the importance of sequential decision-making is established by RL-Extract performing significantly better than the meta-classifier (7.0\% on \emph{Shootings} over all entities). This is also due to the fact that the meta-classifier aggregates over the entire set of extra documents, including the long tail of noisy, irrelevant documents.
 Finally, we see the advantage of enabling the RL system to select queries as our full model RL-Extract obtains significant improvements over RL-Basic on both domains. The full model also outperforms RL-Query, demonstrating the importance of performing both query selection and reconciliation in a joint fashion.


Figure~\ref{fig:reward} shows the learning curve of the agent by measuring reward on the test set after each training epoch. The reward improves gradually and the accuracy on each entity increases simultaneously. Table~\ref{table:outputs} provides some examples where our model is able to extract the right values when the baseline fails. One can see that in most cases this is due to the model making use of articles with prototypical language or articles containing the entities in readily extractable form.

 \begin{table*}[t]
\centering
\begin{tabular}{ c | c | c || c | c | c | c | c  } 
\textbf{Reconciliation} & \multirow{ 2}{*}{\textbf{Context}} & \multirow{ 2}{*}{\textbf{Reward}} &   \multicolumn{4} { c | }{\textbf{Accuracy}} &  \multirow{ 2}{*}{\textbf{Steps}}\\ \cline{4-7}
\textbf{(RL-Extract)} & &  & \textbf{S} & \textbf{K} & \textbf{W} & \textbf{C} & \\ \hline 
\emph{Confidence} & tf-idf & Step & 47.5 & 71.5 & 70.4 & 60.1 & 8.4 \\ 
\emph{Majority} & tf-idf & Step & 43.6 & 71.8 & 69.0 & 59.2 & 9.9 \\ \hline
Replace & \emph{No context} & Step & 44.4 & 77.1 & 72.5 & 63.4 & 8.0 \\ 
Replace & \emph{Unigram} & Step &  48.9 & 76.8 & 74.0 & 63.2 & 10.0 \\ \hline
Replace & tf-idf & \emph{Episode} & 42.6 & 62.3 & 68.9 & 52.7 & 6.8 \\ \hline \hline
Replace & tf-idf & Step & \textbf{50.0} & \textbf{77.6} & \textbf{74.6} & \textbf{65.6} & 9.4\\ 
\end{tabular}
\caption{Effect of using different reconciliation schemes, context-vectors, and rewards in our RL framework (\emph{Shootings} domain). The last row is the overall best scheme  (deviations from this are in \emph{italics}). Context refers to the type of word counts used in the state vector to represent entity context. Rewards are either per step or per episode. (\textbf{S}:~ShooterName, \textbf{K}:~NumKilled, \textbf{W}:~NumWounded, \textbf{C}:~City, \textbf{Steps}:~Average number of steps per episode)}
\label{table:results2}
\end{table*}

\paragraph{Analysis} We also analyze the importance of different reconciliation schemes, rewards and context-vectors in RL-Extract on the Shootings domain (Table~\ref{table:results2}). In addition to simple replacement (Replace), we also experiment with using Confidence and Majority-based reconciliation schemes for RL-Extract. We observe that the Replace scheme performs much better than the others (2-6\% on all entities) and believe this is because it provides the agent with more flexibility in choosing the final values.

From the same table, we see that using the tf-idf counts of context words as part of the state provides better performance than using no context or using simple unigram counts. In terms of reward structure, providing rewards after each step is empirically found to be significantly better (>10\% on average) compared to a single delayed reward per episode. 
The last column shows the average number of steps per episode -- the values range from 6.8 to 10.0 steps for the different schemes. The best system (RL-Extract with Replace, tf-idf and step-based rewards) uses 9.4 steps per episode.

\section{Conclusions}

In this paper, we explore the task of acquiring and incorporating external evidence to improve information extraction accuracy for domains with limited access to training data. This process comprises issuing search queries, extraction from new sources and reconciliation of extracted values, repeated until sufficient evidence is obtained. We use a reinforcement learning framework and learn optimal action sequences to maximize extraction accuracy while penalizing extra effort. We show that our model, trained as a deep Q-network, outperforms traditional extractors by 7.2\% and 5\% on average on two different domains, respectively. We also demonstrate the importance of sequential decision-making by comparing our model to a meta-classifier operating on the same space, obtaining up to a 7\% gain. 


\section*{Acknowledgements}
We thank David Alvarez, Tao Lei and Ramya Ramakrishnan for helpful discussions and feedback, and the members of the MIT NLP group and the anonymous reviewers for their insightful comments. We also gratefully acknowledge support from a Google faculty award.

\bibliography{references}
\bibliographystyle{emnlp2016}

\end{document}


\maketitle

\section{Framework and Model}
\begin{figure}[!hbp]
  \includegraphics[width=0.9\linewidth]{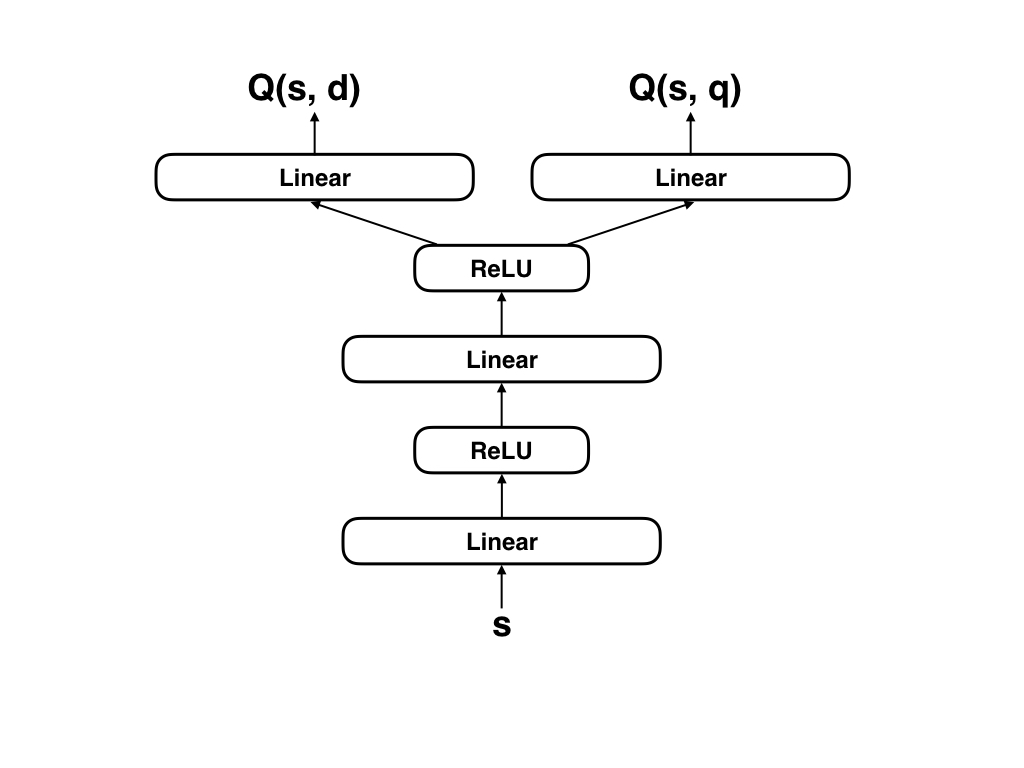}
\caption{DQN architecture}
	\label{fig:model}
\end{figure}

Table~\ref{table:features} lists examples of features used in our Maxent classifier. Figure~\ref{fig:model} illustrates the architecture of our DQN model.
\begin{table}[!htbp]
\centering
\begin{tabular}{| l |} \hline
\textbf{Features:} \\ \hline
isMaleName, isFemaleName \\ \hline
isCapital, isLongWord, isShortWord \\ \hline
isDigit, containsDigit, isNumberWord \\ 
isOrdinalWord \\ \hline
isFullCity, isPartialCity \\ \hline
\end{tabular}
\caption{Examples of unigram features used in Maximum Entropy classifier. The same features are also calculated for neighboring words in the surrounding context. }
\label{table:features}
\end{table}

\bibliography{references}
\bibliographystyle{acl}